\PassOptionsToPackage{usenames,dvipsnames,table,xcdraw}{xcolor}
\documentclass[10pt, a4paper]{article}

\usepackage[]{lrec-coling2024} 

\usepackage{soul}
\usepackage{color}
\usepackage{amsfonts,amssymb,mathrsfs}
\usepackage[usenames,dvipsnames,table,xcdraw]{xcolor}
\usepackage{algorithm}
\usepackage[noend]{algpseudocode}

\usepackage{bm}
\usepackage{amsmath}
\usepackage{makecell}

\usepackage{latexsym}
\usepackage{multirow}

\usepackage{float}

\usepackage{booktabs}
\usepackage{multirow}
\usepackage{graphicx}

\usepackage{graphicx}
\usepackage{lipsum}
\usepackage{relsize}
\sethlcolor{cyan}  

\title{\textbf{Semantic Role Labeling Guided Out-of-distribution Detection}}

\name{\parbox{\textwidth}{\centering Jinan Zou\textsuperscript{1,\dag} \thanks{\noindent\textsuperscript{\dag}  Indicates equal contribution as first authors}, Maihao Guo\textsuperscript{1,\dag}, Yu Tian\textsuperscript{2,\dag}, Yuhao Lin\textsuperscript{1}, Haiyao Cao\textsuperscript{1}, Lingqiao Liu\textsuperscript{1},  Ehsan Abbasnejad\textsuperscript{1}, Javen Qinfeng Shi\textsuperscript{1,*} \thanks{\noindent\textsuperscript{*} Corresponding author}}}

\address{\textsuperscript{1} Australian Institute for Machine Learning, University of Adelaide, Adelaide, Australia \\
         \textsuperscript{2}Harvard University, Cambridge, USA \\
         \{jinan.zou, javen.shi\}@adelaide.edu.au, ytian11@meei.harvard.edu}

\abstract{
Identifying unexpected domain-shifted instances in natural language processing is crucial in real-world applications. Previous works identify the out-of-distribution (OOD) instance by leveraging a single global feature embedding to represent the sentence, which cannot characterize subtle OOD patterns well. Another major challenge current OOD methods face is learning effective low-dimensional sentence representations to identify the hard OOD instances that are semantically similar to the in-distribution (ID) data.
In this paper, we propose a new unsupervised OOD detection method, namely Semantic Role Labeling Guided Out-of-distribution Detection (SRLOOD), that separates, extracts, and learns the semantic role labeling (SRL) guided fine-grained local feature representations from different arguments of a sentence and the global feature representations of the full sentence using a margin-based contrastive loss. A novel self-supervised approach is also introduced to enhance such global-local feature learning by predicting the SRL extracted role. The resulting model achieves SOTA performance on four OOD benchmarks, indicating the effectiveness of our approach. The code is publicly accessible
via \url{https://github.com/cytai/SRLOOD}. 
\\ \newline \Keywords{Out-of-distribution Detection, Semantic Role Labeling, Domain Shift} }

\begin{document}

\maketitleabstract

\section{Introduction}
Recent advances in natural language processing have shown tremendous improvements in various natural language classification tasks. Natural language classification is usually formulated as a close-set problem, where training and testing samples are from the same domain/distribution. 
Despite the accurate predictions on the inlier close-set classes, the classifier often fails to properly identify out-of-distribution (OOD) instances from other unknown/unexpected domains that deviate from the close-set training distribution, bringing risks to real-world scenarios. 
Tackling such failure cases is crucial to real-world safety-critical NLP applications. For instance, OOD instances can be represented by unknown sentences from different domains or distributions, such as semantically shifted sentences that can be incorrectly
predicted as a part of the inlier classes, leading to potential impairment to user trust~\cite{arora2021types}. 
Although large language models (LLMs) are revolutionizing the field of NLP, they are prone to OOD and even adversarial inputs~\cite{robustness}. OOD detection can be applied to directly handle OOD inputs and avoid potentially harmful responses ~\cite{harmless}. 

Despite the importance, little literature has addressed the problem of OOD detection in NLP. One proposed method is to train a model to increase the inter-class discrepancy of in-distribution (ID) classes and tends to depend on classification uncertainty or latent embedding distance to detect OOD instances \cite{zhou-etal-2021-contrastive}. 
The high classification uncertainty association with OOD instances (i.e., max softmax or energy) is intuitive, but it obtains a few caveats.  
One of the major issues is that classification uncertainty happens when samples are close to classification decision boundaries. However, there is no guarantee that all OOD instances will be close to classification boundaries (i.e., subtle OOD samples may share similar semantic features to ID data), leading to subpar performance in detecting OOD samples.  Moreover, complicated inlier sentences containing more outlier components, such as punctuation and discourse fillers, tend to fall close to the decision boundary, 
which can incorrectly lead to high classification uncertainty. 
Latent embedding-based approaches rely on the assumption that the OOD instance resides outside a bounded or unbounded latent hyperspace constructed by the ID feature distributions~\cite{zhou-etal-2021-contrastive,hendrycks-etal-2020-pretrained,cao2022deep,DBLP:journals/corr/abs-2111-00506, tan-etal-2019-domain}. However, it is challenging to define such a latent hyperspace to encode all possible ID features, significantly affected by many outlier components from a sentence, and the aforementioned subtle OOD issue still exists.

\begin{figure*}[ht]
\begin{center}
\resizebox{0.9\linewidth}{!}{%
\includegraphics[width=1\linewidth]{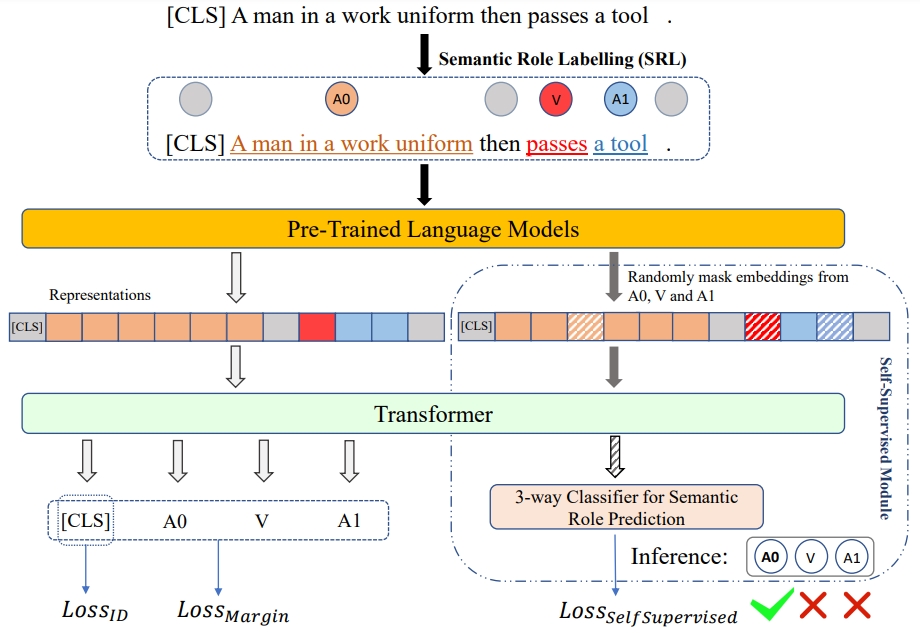}}
\end{center}
\vspace{-.15in}
\caption{Model architecture of our framework. The Transformers including the pre-traiend language model and a subsequent encoder are guided by SRL, extracting global and local representations of input sequence according to the semantic roles A0, V, A1, and masking them according to the semantic roles A0, V, A1 to construct the Self-Supervised Module. An additional 3-way Classifiers take the Transformer representation of A0, V or A1 MASKs as input, and predict their semantic roles.}\vspace{-.15in}
\label{yyds}
\end{figure*}

In this paper, we propose a new OOD detection method designed for NLP tasks, namely Semantic Role Labeling Guided Out-of-distribution Detection (SRLOOD), simultaneously extracting, separating, and learning both global and SRL-guided local fine-grained feature representations through a margin-based contrastive loss and self-supervision. We identify a critical shortcoming in current NLP models' ability to detect OOD cases: they lack nuanced, low-dimensional local representations that are crucial for identifying OOD instances semantically similar to ID data. SRL emerges as a key technique, that aims at extracting vital local features while effectively omitting outlier elements, including punctuation and discourse fillers.
This innovative integration of SRL into our methodology leads to a remarkable enhancement in both the performance and efficiency of OOD detection. 
In particular, our contributions can be summarised into three folds: 
\begin{itemize}

\item 
We propose SRLOOD that learns fine-grained low-dimensional representations by increasing the inter-class discrepancies between the concatenation of the global and SRL-guided local features of different ID classes. Our proposed SRLOOD aims to effectively eliminate the outlier phrases (e.g., punctuation and discourse fillers) and extract key local semantic components (e.g., verbs and arguments) from a sentence to better characterize subtle OOD samples;

\item A novel self-supervised pretext task is also proposed to strengthen the relations between different local arguments, further facilitating the optimization of SRL-guided local features; and

\item A Transformer block is introduced to resemble some of the SRL-guided features from a sentence, so our model is enforced to learn discriminative representations through such strong perturbations for the better discriminability of subtle semantic features. 

\item Extensive experiments on four different OOD benchmarks show that our resulting model achieves the best performance on four different scoring functions. 
\end{itemize}

\section{Related Work}\textbf{Out-of-distribution detection}
Machine learning aims to design models that can learn generalizable knowledge from training data. The success of machine learning models lies in the assumption that training and test data share the same distribution. However, in many real-world tasks, it is unknown whether the training and test data share the same distribution. For online LLMs such as ChatGPT that interact with users, the inputs out of the training distribution is prevalent\cite{robustness}. This potential distribution gap is known as OOD and can be a major issue, with the performance of classical ML models often deteriorating. To handle the OOD issue, OOD detection aims to detect whether test data is from the training distribution. Based on the availability of OOD data, recent methods can be categorized into classification methods, density-based methods, and distance-based methods \cite{yang2021generalized}. Classification methods often formulate the OOD task as a one-class classification problem, then use appropriate methods to solve it \cite{ruff2018deep,chen2021deep,tian2021pixel,hendrycks2018deep,lee2018training,dhamija2018reducing,morteza2022provable,chen2022deep}. 
\citet{hendrycks17baseline} proposed a softmax prediction probability baseline for error and out-of-distribution detection across several architectures and numerous datasets. Density-based methods \cite{cao2022deep,abati2019latent,zisselman2020deep,kirichenko2020normalizing} in OOD detection explicitly model
the in-distribution with some probabilistic models and flag test data in low-density regions as OOD. \citet{zong2018deep} utilizes a deep autoencoder to generate a low-dimensional representation and reconstruction error for each input data point, which is further fed into a Gaussian Mixture Model for anomaly detection. The main idea of distance-based methods is that the testing
OOD samples should be relatively far away from the
centroids of in-distribution classes \cite{lee2018training,chen2020boundary,van2020uncertainty,Zaeemzadeh_2021_CVPR}. Previous methods primarily studied for computer vision \cite{Lin_2021_CVPR,Huang_2021_CVPR,Zaeemzadeh_2021_CVPR,zhou2022rethinking,dong2022neural} and OOD detection has been overlooked in NLP. Only few works recently that adapted the solutions designed for images into the text to leverage the features representation of an entire sentence for detecting the OOD case. For example, \citet{zhou-etal-2021-contrastive} adapted a contrastive OOD detection from computer vision using a pre-trained Transformer to improve the compact news of representations and evaluate the trained classifier on the four text datasets. \\
In contrast, we propose a self-supervised SRL method to learn fine-grained feature representations of text data and shows
that is a surprisingly effective approach for OOD detection.\\

\noindent\textbf{Semantic Role Labeling}
Semantic role labelling(SRL) leads to the advancement of many NLP tasks and applications due to the clear detection of augments regarding predicates. For example, \citet{sarzynska2021detecting} proposed a BERT-based model incorporating semantic role labelling, which significantly improves the text understanding ability of the model. \citet{chen2021human} used the verb-specific semantic role, a variant of semantic role labelling, for the controllable image captioning, which is a task about image description. Conditioned on the semantic role representation. More recently,  \citet{ross-etal-2022-tailor} proposed a Tailor model for the sequence-to-sequence task, which gained a great improvement in measuring the reliance on syntactic heuristics. \\

\noindent\textbf{Self-supervised Learning}
Self-supervised learning method and has been soaring and achieving big success in representative learning because of the powerful generalization ability. BERT (Pre-training of deep bidirectional Transformers for language understanding) proposed by \citet{devlin2018bert} are fine-tuned for many downstream tasks. as a result, BERT has become a milestone of not only NLP but also the development of self-supervised learning.
\citet{baevski2022data2vec} built a platform based on a self-supervised method for either speech, text or computer vision. Previous research has shown that the self-supervised method drastically improves the OOD detection performance on the difficult near-domain outliers~\citet{hendrycks2019using,tian2023self}. Self-supervised learning methods tackle the OOD in two aspects: (1) the enhancement of
feature quality can improve OOD performance; (2) some well-designed surrogate tasks can help reveal the anomalies from
OOD samples\cite{yang2021generalized}.

\section{Methodology}

Generally, the OOD instances can be defined as instances $(x,y)$ sampled from an underlying distribution other than the training in-distribution  $P(\mathcal{X}_{train},\mathcal{Y}_{train})$, where $\mathcal{X}_{train}$ and $\mathcal{Y}_{train}$ are the training corpus and training label set. Specifically, an instance $(\boldsymbol{x}, y)$ is primarily deemed OOD if $y\notin\mathcal{Y}_{train}$ to be consistent with previous works~\cite{hendrycks17baseline,hendrycks2018deep,hendrycks-etal-2020-pretrained,zhou-etal-2021-contrastive}. Following the previous work \cite{zhou-etal-2021-contrastive}, we formally define the OOD detection task. Given the main task of natural language classification, the OOD detection task is the binary classification of each instance $\boldsymbol{x}$ as either ID or OOD, judged by its OOD score computed with scoring function $f(x)\rightarrow\mathbb{R}$. A lower OOD score value indicates ID where $y\in\mathcal{Y}_{train}$ and a higher OOD score value indicates OOD where $y\notin\mathcal{Y}_{train}$ ($y$ is the underlying label for $\boldsymbol{x}$ and is unknown at inference).

The key idea of our proposed model, SRLOOD, is extracting and learning the SRL-guided fine-grained local representation. Building on top of this representation, a novel supervised approach is introduced to enhance such local argument representation.

Our framework consists of (1) semantic role labelling, (2) an SRL-guided self-supervised module, and (3) OOD detection with OOD scoring functions as illustrated in Figure \ref{yyds}.

\begin{algorithm}[!t]
    \caption{Learning Process}
    \label{algo::main}
    \small
    \textbf{Input} ID training set $\mathcal{D}_\text{train}$ and ID validation set $\mathcal{D}_\text{val}$. \\
    \textbf{Output} A trained classifier and an OOD detector. \\
    
    Load the Pre-Trained Transformer and initialize the subsequent Transformer.
    
    \begin{algorithmic}
        \For{$t=1...T$}
            \State Sample a batch from $\mathcal{D}_\text{train}$.
            \State Calculate the ID classification loss $\mathcal{L}_\text{ID}$.
            \State Calculate the contrastive loss $\mathcal{L}_\text{margin}$.
            \State Calculate the self-supervised loss $\mathcal{L}_\text{SSL}$.
            \State $\mathcal{L}_\text{total} = \alpha_{1}\mathcal{L}_\text{ID} + \alpha_{2}\mathcal{L}_\text{Margin} + \alpha_{3}\mathcal{L}_\text{SSL}$.
            \State Update model parameters w.r.t. $\mathcal{L}_\text{total}$.
            \State \textbf{if} $t \;\%\; \text{evaluation steps} = 0$, \textbf{then}:
            \State \hspace{15pt}Fit the OOD detector on $\mathcal{D}_\text{val}$.
            \State \hspace{15pt}Evaluate both the classifier and OOD detector on $\mathcal{D}_\text{val}$.
        \EndFor
        \State Return the best model checkpoint.
    \end{algorithmic}
\end{algorithm}

\subsection{Semantic Role Labeling}

The task of SRL is to determine the underlying predictive argument structure of a sentence and to provide representations that can answer the basic questions about the meaning of the sentence, including who did what to whom \cite{10.1162/coli.2008.34.2.145}. Therefore the SRL primarily extracts the essential features and passingly filters out outlier phrases (e.g., punctuation and discourse fillers).
We leverage off-the-shelf SRL-BERT~ \cite{SRLBERT} to label each token sequence in a batch with Propbank~ \cite{kingsbury2003propbank} semantic roles proto-agent, verb, and proto-patient, then labeled tokens are recorded into sets A0, V, and A1 respectively, as illustrated in Figure \ref{yyds}. Each token sequence is fed to the pre-trained language model, whose output is fed to the Transformer block. We compute the mean of A0, V, A1 embeddings $\boldsymbol{\mu}_{A0}$, $\boldsymbol{\mu}_{V}$, and $\boldsymbol{\mu}_{A1}$ pooled from the Transformer's output for fine-grained feature representations.

\subsection{Self Supervised Learning based on SRL}

Our proposed SRLOOD framework uses SRL to extract and learn key local semantic features and use self-supervision to further strengthen such fine-grained local representations. 
We introduce strong perturbation by randomly masking a certain percentage of SRL-extracted local representations for better generalization on detecting hard OOD instances. 

Guided by SRL, strong perturbation is independently exerted on the pre-trained language model representations of A0, V, and A1 according to a generated and recorded supervising ground truth label for each sequence. The perturbed embeddings are input to the Transformer encoder. Subsequently, we compute the mean embeddings of A0, V, or A1 from the Transformer's output and use them for an auxiliary three-way classification task. This task aims to improve the feature discriminability by predicting the semantic role of a given embedding,  computing the mean embeddings, and inputting one of them to a classifier for semantic role prediction according to its self-supervising label. To this end, our framework is consisted of the pre-trained language model, the Transformer head, the SRL-guided pooling, and the 3-way self-supervised classifier. This framework is optimized by the loss functions introduced in the next section.

\subsection{Loss Functions}

We adopt the margin-based contrastive loss that drives the model to encode tokens in the same ID class with adjacent SRL-guided comprehensive representation measured by L2 distances:
%
%
\vspace{-.15in}
\begin{equation}
\begin{aligned}
        \mathcal L_\text{margin} =\frac{1}{m d} \left[ \sum_{i=1}^{m} \frac{1}{|P(i)|} \sum_{p \in P(i)}
        \left(\left\|\boldsymbol{h}_{i}-\boldsymbol{h}_{p}\right\|^{2} \right) \right.\\
        \left.+ \sum_{i=1}^{m} \frac{1}{|N(i)|} \sum_{n \in N(i)}\left(\xi-\left\|\boldsymbol{h}_{i}-\boldsymbol{h}_{n}\right\|^{2}\right)_{+} \right]
\end{aligned}
\end{equation}

where $P(i)$ is the subset of training data with the same class as instance $i$, $N(i)$ is the subset of training data with different class labels from instance $i$, $m$ is the number of instances in the entire training set. The $d$ is the dimensionality of comprehensive representation of a sequence $\boldsymbol{h} = Concat(\boldsymbol{h}_{[CLS]}; \boldsymbol{\mu}_{A0}; \boldsymbol{\mu}_{V}; \boldsymbol{\mu}_{A0})$, where $\boldsymbol{h}_{[CLS]}$ is the [CLS] embedding, $\boldsymbol{\mu}_{A0}$, $\boldsymbol{\mu}_{V}$, $\boldsymbol{\mu}_{A1}$ are the mean embeddings pooled from the Transformer's output according to A0, V, A1 respectively. The Margin Loss will give rise to clusters in the latent space of $\boldsymbol{h}$. Combined with cross-entropy losses $\mathcal L_\text{ID}$ and $\mathcal L_\text{SSL}$ from the ID sequence classification task and the self-supervised task, respectively. The total loss is their weighted sum with hyper-parameters $\alpha_{1}$, $\alpha_{2}$ and $\alpha_{3}$:

\begin{equation}
    \begin{aligned}
        \mathcal L_\text{total} =\alpha_{1}\mathcal L_\text{ID}+\alpha_{2}\mathcal L_\text{margin}+\alpha_{3}\mathcal L_\text{SSL} .
    \end{aligned}
\end{equation}

\subsection{Scoring Functions}
During OOD inference, we extract the local key components and features using SRL-BERT~\cite{SRLBERT} based on a previously fine-tuned language model backbone. We compute the mean embeddings $\boldsymbol{\mu}_{A0}$, $\boldsymbol{\mu}_{V}$, and $\boldsymbol{\mu}_{A1}$  for local feature representations. The [CLS] embedding $\boldsymbol{h}_{[CLS]}$ is used for global feature representations. The global and local representations are then concatenated together to produce the final feature vector to represent a sentence 
\begin{equation}
    \begin{aligned}
        \boldsymbol{h}= Concat(\boldsymbol{h}_{[CLS]}; \boldsymbol{\mu}_{A0}; \boldsymbol{\mu}_{V}; \boldsymbol{\mu}_{A0}) . 
    \end{aligned}
\end{equation}

For a fair comparison, we use the same OOD scoring functions as \citet{zhou-etal-2021-contrastive}. For the validation set $\mathcal{D}^{val} = \{ (\mathbf{x}_i,\mathbf{y}_i) \}_{i=1}^{|\mathcal{D}^{val}|}$, we computed the Mahalanobis distance based on the class mean embedding $\boldsymbol{\mu}_{c} =\mathbb{E}_{y_{i}=c}\left[\boldsymbol{h}_{i}\right]$, $c\in C$ the number of classes, and its covariance $\boldsymbol{\Sigma} =\mathbb{E}\left[\left(\boldsymbol{h}_{i}-\boldsymbol{\mu}_{y_{i}}\right)\left(\boldsymbol{h}_{i}-\boldsymbol{\mu}_{y_{i}}\right)^{\intercal}\right]$, where $i = 1,...,C$. The OOD score $S$ is then defined as the minimum Mahalanobis distance among the $C$ ID classes given an instance $\mathbf{x}$ during inference:

\begin{equation}
    \begin{aligned}
        S=-\min_{c=1}^{C}(\boldsymbol{h}-\boldsymbol{\mu}_{c})^\intercal\boldsymbol{\Sigma}^{\dagger}(\boldsymbol{h}-\boldsymbol{\mu}_{c}), 
    \end{aligned}
\end{equation}
where $\boldsymbol{\Sigma}^{\dagger}$ denotes the pseudo-inverse of the covariance matrix $\boldsymbol{\Sigma}$. Such a distance considers both the global sentence features and the SRL-guided local features, enabling better performance on OOD detection. 

For cosine similarity, we compute the maximum cosine similarity of the concatenated feature representation $\boldsymbol{h}$ to instance features of the validation set $\mathcal{H}^{val} = \{ (\boldsymbol{h}_{i},\mathbf{y}_i) \}_{i=1}^{|\mathcal{H}^{val}|}$. The OOD score is computed as 

\begin{equation}
    \begin{aligned}
        S=-\max_{i=1}^{|\mathcal{H}^{val}|} \cos \left(\boldsymbol{h}, \boldsymbol{h}_{i} \right).
    \end{aligned}
\end{equation}

Maximum Softmax Probability (MSP)~\cite{hendrycks17baseline} and Energy Score (Energy) \cite{NEURIPS2020_f5496252} represent the class of probabilistic scoring functions. Although the MSP is biased, not aligned with the density of the inputs~\cite{NEURIPS2020_f5496252}, it is widely adopted as a baseline for OOD detection. For $C$ training classes in the softmax layer, the MSP score is defined by the  maximum class probability:

\begin{equation}
    \begin{aligned}
        S = 1 - \max_{j=1}^C \bm{p}_j.
    \end{aligned}
\end{equation}


\citet{NEURIPS2020_f5496252} estimates the probability density of inputs as:
\begin{equation}
    \begin{aligned}
        S = -\log \sum_{j=1}^C \exp(\bm{w}_j^\intercal \boldsymbol{h}_{softmax}),
    \end{aligned}
\end{equation}
where $\bm{w}_j \in \mathbb{R}^{d}$ is the weight of the $j^{th}$ class in the softmax layer, $\boldsymbol{h}_{softmax}$ is the input to the softmax layer.
A higher energy score S indicates a greater likelihood of being OOD data, thereby suggesting a lower likelihood of being ID data. 

\subsection{Datasets}
Previous studies on OOD detection mostly focus on computer vision, while few have been made on natural language processing.
\citet{zhou-etal-2021-contrastive} propose a extensive benchmarks for OOD detection on natural language processing and use different pairs of NLP datasets as ID and OOD data.  Following \citet{zhou-etal-2021-contrastive}, we use the same NLP datasets and same criterion on choosing ID and OOD data to evaluate our proposed method.  The ID datasets correspond to three categories of natural language classification tasks as following: 
\begin{itemize}
    \item \textbf{Sentiment Analysis}
Following \cite{zhou-etal-2021-contrastive}, we use SST2\cite{socher2013recursive} and IMDB \cite{maas-etal-2011-learning} as our ID datasets, which are both sentiment analysis datasets. Note that both datasets belong to the same task and are not condisered OOD to each other.
 \end{itemize}

\begin{itemize}
    \item \textbf{Topic Classification}
Following \cite{zhou-etal-2021-contrastive}, we use 20 Newsgroup dataset  \cite{lang1995newsweeder} as our ID dataset, which is a dataset for topic classification containing 20 classes.
\end{itemize}

\begin{itemize}
    \item \textbf{Question Classification}
Following \cite{zhou-etal-2021-contrastive}, we use TREC-10 dataset\cite{li2002learning} as our ID dataset, which classifies questions based on the types of their sought-after answers.
\end{itemize}

Moreover, for the above three tasks, any pair of datasets for different tasks can be regarded as OOD to each other. Besides, following \citet{zhou-etal-2021-contrastive}, we employ for additional datasets solely as the OOD data: concatenations of the premises and respective hypotheses from two \textbf{NLI} datasets RTE \cite{dagan2005pascal,haim2006second, giampiccolo2007third,bentivogli2009fifth} and MNLI\cite{williams-etal-2018-broad}, the English source side of \textbf{Machine Translation}(\textbf{MT}) datasets English-German WMT16 \cite{bojar2016findings} and Multi30K \cite{elliott-etal-2016-multi30k}.

\subsection{Evaluation Metrics}
We adopted the same two metrics \cite{zhou-etal-2021-contrastive} commonly used for measuring OOD detection performance in machine learning researches \cite{hendrycks17baseline,lee2018training}:  \textbf{AUROC} and \textbf{FAR95}. \textbf{AUROC} is the area under the receiver operating characteristic (ROC) curve. It compares the true positive rate (TPR) to the false positive rate (FPR). \textbf{FAR95} is the probability of mistakenly classifying OOD as ID at a 95\% TPR.\\
\subsection{Experiments Details}
We conducted all experiments based on the same codebase and used the same ${\rm RoBERTa}_{\rm LARGE}$ from previous work \cite{zhou-etal-2021-contrastive}. The Transformer encoder has $3$ layers and $16$ attention heads. The weights $\alpha_{1}=1$, $\alpha_{2}=3$, $\alpha_{3}=1$. The warm-up ratio for learning rate is $0.06$. 
The batch size is $12$. 
We use AdamW \cite{Adam} to optimized our model, and a learning rate of $1e-5$ and weighted decay $0.01$.
We pick the masking probability that optimize the average OOD detection performance, $30\%$ for SST2 and IMDB, and $50\%$ for TREC-10 and 20NG, guided by Figure  \ref{maskingscore}.
The model is trained for $10$ epochs with runtime ranging from $5$ hours to $10$ hours on one Tesla V100 GPU. We further discuss the performance of taking different masking probabilities in Figure \ref{maskingscore}.  
Please note that we manually select all hyper-parameters based on the AUC and FAR performance on testing sets. 
The total number of model parameters is $392$M.  All the hyper-parameters are tuned on the development sets.
\\
\textbf{Compared Methods.} We compare our method with three baselines: OOD detection using probabilities from softmax distributions w/o $\mathcal L_\text{Cont}$-MSP \cite{hendrycks17baseline}, fine-tuning the Transformers with supervised contrastive loss w/ $\mathcal L_\text{SCL}$, and with  margin-based loss w/ $\mathcal L_\text{margin}$ 
 \cite{zhou-etal-2021-contrastive}. 


\subsection{Main Results}

\begin{table*}[ht!]
\centering
\tiny
\resizebox{\textwidth}{!}{%
\begin{tabular}{ccccccc}
\hline
\multicolumn{2}{l}{\textbf{AUROC$\uparrow$ /FAR95$\downarrow$}}& \textbf{Avg} & \textbf{SST2} & \textbf{IMDB} & \textbf{TREC-10} & \textbf{20NG} \\ \hline
\multirow{4}{*}{\makecell[c]{w/o $\mathcal L_\text{Cont}$\\ \cite{zhou-etal-2021-contrastive}}} & MSP & 94.1/35.0 & 88.9/61.3 & 94.7/40.6 & 98.1/7.6 & 94.6/30.5 \\
 & Energy & 94.0/34.7 & 87.7/63.2 & 93.9/49.5 & 98.0/10.4 & 96.5/15.8 \\
 & Maha & 98.5/7.3 & 96.9/18.3 & 99.8/0.7 & 99.0/2.7 & 98.3/7.3 \\
 & Cosine & 98.2/9.7 & 96.2/23.6 & 99.4/2.1 & 99.2/2.3 & 97.8/10.7 \\ \hline
\multirow{4}{*}{\makecell[c]{w/ $\mathcal L_\text{SCl}$\\ \cite{zhou-etal-2021-contrastive}}} & $\mathcal L_\text{scl}$+MSP & 90.4/46.3 & 89.7/59.9 & 93.5/48.6 & 90.2/36.4 & 88.1/39.2 \\
 & $\mathcal L_\text{scl}$+Energy & 90.5/43.5 & 88.5/64.7 & 92.8/50.4 & 90.3/32.2 & 90.2/26.8 \\
 & $\mathcal L_\text{scl}$+Maha & 98.3/10.5 & 96.4/26.6 & 99.6/2.0 & 99.2/1.9 & 97.9/11.6 \\
 & $\mathcal L_\text{scl}$+Cosine & 97.7/13.0 & 95.9/28.2 & 99.2/4.2 & 99.0/2.4 & 96.8/17.0 \\ \hline
\multirow{4}{*}{\makecell[c]{w/ $\mathcal L_\text{margin}$\\ \cite{zhou-etal-2021-contrastive}}} & $\mathcal L_\text{margin}$+MSP & 93.0/33.7 & 89.7/49.2 & 93.9/46.3 & 97.6/6.5 & 90.9/32.6 \\
 & $\mathcal L_\text{margin}$+Energy & 93.9/31.0 & 89.6/48.8 & 93.4/52.1 & 98.4/4.6 & 94.1/18.6 \\
 & $\mathcal L_\text{margin}$+Maha & 99.5/1.7 & 99.9/0.6 & 100/0 & 99.3/0.4 & 98.9/6.0 \\
 & $\mathcal L_\text{margin}$+Cosine & 99.0/3.8 & 99.6/1.7 & 99.9/0.2 & 99.0/1.5 & 97.4/11.8 \\ \hline
\multirow{4}{*}{\textbf{Ours}} & MSP & \hl{94.8/24.7}  &  \hl{90.8/46.4}    &   \hl{97.0/18.3}   &   \hl{98.6/2.5}      &    92.9/31.4 \\
 & Energy & \hl{95.7/20.7} & \hl{90.4/45.5} & \hl{97.0/19.9} & \hl{98.5/3.2} & \hl{96.9/14.0} \\
 & Maha & \hl{99.6/0.8} & 99.4/2.2 & 99.5/0.7 & \hl{ 99.9/0} & \hl{99.1/0.8} \\
 & Cosine & \hl{99.0/3.5} & 98.7/6.5 & 98.7/4.8 & \hl{99.5/0.4} & \hl{98.9/2.3}
 \\\hline
\end{tabular}%
}
\caption{OOD Detection performance (in \%) of ${\rm RoBERTa}_{\rm LARGE}$ trained on the four different ID datasets. 
Following the highlight standard \cite{zhou-etal-2021-contrastive}, the results of our SRL-guided Self Supervision method achieving SOTA on both evaluation metrics are highlighted in  \hl{blue}.}
\label{mainresult}
\end{table*}

As demonstrated in Table \ref{mainresult}, our SRLOOD model excels in performance compared to three diverse State-of-the-Art (SOTA) techniques across a variety of Out-of-Distribution (OOD) benchmarks. Regardless of the scoring function applied, our model delivers superior Mean Area Under the Curve (AUC) and False Acceptance Rate (FAR) results. This high performance pertains to MSP, energy, Cosine, and Mahalanobis distance scoring.

Our model, even without fine-tuning of classification logits, offers significant advancements over preceding SOTA methodologies. All reported results represent an average of five independent runs, each initiated with different random seeds. Among the various OOD detection functions, Mahalanobis and cosine distances prove to be the most effective, outstripping the MSP and energy baseline by a significant margin, as documented in~\cite{hendrycks17baseline}. The reason behind the superior performance of Mahalanobis and cosine distances in detecting distributional differences can be attributed to their enhanced capacity to encapsulate such differences. 

When contrasted with baseline methodologies that do not employ contrastive loss, our approach notably improves the FAR by an average of 10\%, 14\%, 7\%, and 6\% for MSP, energy, Mahalanobis, and cosine distances respectively. When pitted against earlier SOTA techniques~\cite{zhou-etal-2021-contrastive} that use $\mathcal L_\text{SCl}$, $\mathcal L_\text{margin}$ losses, our strategy significantly surpasses them in terms of mean AUC and FAR across all four OOD distance measures. To be precise, our model furnishes a minimum of 9\%, 11\%, 1\%, and 0.3\% enhancements in mean FAR when utilizing MSP, energy, Mahalanobis, and cosine distances, respectively. This consistent improvement showcases the efficiency of our model in OOD detection across various distance measurements and datasets.

In terms of models trained on different In-Distribution (ID) datasets, our model, when applying Mahalanobis and cosine distances, achieves near-perfect OOD detection on SST2, IMDB, and TREC-10 datasets, aligning with the previous SOTA's performance with $\mathcal L_\text{margin}$. While the 20 Newsgroup dataset, comprising articles from multiple genres, offered room for improvement when tackled with the previous SOTA method, our approach managed to deliver near-perfect OOD detection on this dataset, surpassing the prior method's performance by a significant margin. It is noteworthy that our model achieves unrivaled performance in both AUROC and FAR95 on all four datasets evaluated.

Furthermore, we assess the performance of our model across all OOD scoring functions. 
This underscores the exceptional ability of our framework to perform optimally across a very diverse range of scoring functions to enhance the OOD performance. 
Comparisons between our approach and the prior SOTA \cite{zhou-etal-2021-contrastive}, as presented in Table~\ref{tab:my-table2} and \ref{tab:my-table3}, illustrate our model's performance on four different ID datasets, measured in terms of AUROC and FAR95, respectively. 
In comparison with the previous state-of-the-art performance, our method successfully lowers the FAR95 on 3 out of 4 distinct ID datasets, among which, falsely accepting OOD as ID of IMDB is drastically alleviated from 24.7\% to 10.9\%. These results also highlight that our method surpasses previous state-of-the-art AUROC on all 4 distinct ID datasets. 

\begin{table}[ht]
\scalebox{0.7}{
\begin{tabular}{lccccc}
\hline
Methods             & \multicolumn{1}{l}{Overall} & \multicolumn{1}{l}{SST2} & \multicolumn{1}{l}{IMDB} & \multicolumn{1}{l}{TREC-10} & \multicolumn{1}{l}{20NG} \\ \hline
\cite{zhou-etal-2021-contrastive} & 96.4                        & 94.7                     & 96.8                     & 98.6                        & 95.3                     \\ \hline
\textbf{Ours}       & \textbf{97.3}               & \textbf{94.8}            & \textbf{98.1}            & \textbf{99.1}               & \textbf{97.0}            \\ \hline
\end{tabular}
}
\caption{Comparisons of overall AUROC$\uparrow$.The results of our SRL-guided Self Supervision method achieving SOTA are highlighted in \textbf{bold}.}
\label{tab:my-table2}
\end{table}

\begin{table}[ht]
\scalebox{0.7}{
\begin{tabular}{lccccc}
\hline
Methods             & \multicolumn{1}{l}{Overall} & \multicolumn{1}{l}{SST2} & \multicolumn{1}{l}{IMDB} & \multicolumn{1}{l}{TREC-10} & \multicolumn{1}{l}{20NG} \\ \hline
\cite{zhou-etal-2021-contrastive} & 17.6&25.1& 24.7& 3.3& 17.3                     \\ \hline
\textbf{Ours}       & \textbf{12.4}&25.2& \textbf{10.9}& \textbf{1.5}& \textbf{12.1}            \\ \hline
\end{tabular}
}
\caption{Comparisons of overall FAR95$\downarrow$.The results of our SRL-guided Self Supervision method achieving SOTA are highlighted in \textbf{bold}.}
\label{tab:my-table3}
\end{table}

\begin{table*}[t!]
\centering
\scriptsize
{\renewcommand{\arraystretch}{1.4}
\resizebox{\textwidth}{15mm}{
\begin{tabular}{ccccccccccccccccc}
\hline
\multirow{2}{*}{\textbf{AUROC}} & \multicolumn{4}{c}{\textbf{SST2}}     & \multicolumn{4}{c}{\textbf{IMDB}}     & \multicolumn{4}{c}{\textbf{TREC-10}}  & \multicolumn{4}{c}{\textbf{20NG}}   \\ 
                       & MSP & Energy & Maha & Cosine & MSP & Energy & Maha & Cosine & MSP & Energy & Maha & Cosine & MSP & Energy & Maha & Cosine \\ \hline
 SST2                   &  -   & -  & -  & -  & -  &  -  &  -  & -  & \textbf{97.8}/96.2    & \textbf{97.3}/96.6     & \textbf{99.8}/98.4    & \textbf{99.1}/97.8       &  \textbf{96.5}/96.3   & \textbf{99.0}/98.1       &  99.2/99.5    &   \textbf{99.0}/99.0     \\
IMDB                    &  -   & -  & -  & -  & -  &  -  &  -  & -  & \textbf{99.5}/99.3    &92.0/99.7        & \textbf{99.7}/99.6     &\textbf{99.6}/99.3       &93.6/94.5     & 96.7/96.9       &\textbf{99.7}/99.0      & \textbf{98.8}/98.4        \\
TREC-10                &\textbf{96.0}/95.1     &\textbf{96.1}/94.9        &\textbf{99.8}/99.5      &\textbf{99.6}/99.0        &\textbf{98.8}/93.8     &\textbf{98.9}/93.3        &99.9/100      &99.8/100        &-     &-        &-      &-        &\textbf{99.3}/88.0     &\textbf{99.9}/92.4        &99.3/99.6      & \textbf{99.5}/96.5       \\
20NG                   &\textbf{96.8}/95.2     &\textbf{97.0}/95.0        &\textbf{100}/100      &99.9/100       &   \textbf{96.5}/95.4  & \textbf{96.6}/95.3       & 99.8/100     & 98.0/99.9       & \textbf{99.6}/99.2    &99.7/99.8        &\textbf{100}/99.8      &\textbf{100}/99.7        &-     &-        &-      & -       \\
MNLI                   &\textbf{83.0}/82.8     &\textbf{82.8}/82.7        &98.4/99.8      &96.6/99.5        &\textbf{95.7}/92.4     &\textbf{95.6}/91.7        &99.8/100      &96.9/99.9        &\textbf{98.0}/96.6     & \textbf{98.0}/97.6       &\textbf{99.8}/99.2      &\textbf{99.0}/98.8        &\textbf{92.1}/91.0     &\textbf{96.8}/94.2        &\textbf{98.9}/98.4      &\textbf{99.1}/97.2        \\
RTE                    &\textbf{89.4}/87.4     &\textbf{88.2}/87.5        &99.9/100      &99.5/99.9       &\textbf{96.1}/92.9     &\textbf{96.0}/92.1        &99.9/100      &98.4/99.9        &\textbf{98.7}/96.6     &\textbf{98.6}/98.1        &\textbf{99.9}/99.6      &\textbf{99.6}/99.2        &\textbf{85.5}/84.5     &\textbf{92.1}/88.7        &\textbf{98.7}/98.2      &\textbf{98.5}/95.6        \\
WMT16                  &\textbf{85.5}/83.9     &\textbf{84.3}/84.0        &98.9/99.9      &97.3/99.4        &\textbf{96.9}/92.9     &\textbf{96.8}/92.2        &99.9/100      &99.8/99.9        &\textbf{97.8}/97.1     &97.8/98.0        &\textbf{99.9}/99.4      &\textbf{99.5}/99.1        &\textbf{91.9}/88.3     &\textbf{96.8}/92.5        & \textbf{99.0}/98.5     & \textbf{98.8}/96.7       \\
Multi30K               &\textbf{94.2}/93.5     &\textbf{93.7}/93.6        &99.5/100      &99.3/99.9        &\textbf{98.1}/95.9     &\textbf{98.3}/95.7        &99.9/100      &99.8/100        &\textbf{99.1}/97.9     &\textbf{99.1}/98.9        &100/99.5      &\textbf{99.9}/99.3        &91.6/93.7     &\textbf{96.9}/96.0        &99.0/99.1      &\textbf{98.6}/98.3        \\ \hline
Avg                    &\textbf{90.8}/89.7     &\textbf{90.4}/89.6        &99.4/99.9      &98.7/99.6        &\textbf{97.0}/93.9     &\textbf{97.0}/93.4       &99.9/100      &98.7/99.9        &\textbf{98.6}/97.6     &\textbf{98.5}/98.4        &\textbf{99.9}/99.3      & \textbf{99.5}/99.0        &\textbf{92.9}/90.9     &\textbf{96.9}/94.1        &\textbf{99.1}/98.9      & \textbf{98.9}/97.4  \\
\bottomrule
\end{tabular}}
\caption{OOD detection AUROC (\%) of ours and w/ $\mathcal L_\text{margin}$ \cite{zhou-etal-2021-contrastive}. The results of our SRL-guided Self Supervision method achieving SOTA on AUROC are highlighted in \textbf{bold}.} 
\label{table3}}
\end{table*}

\begin{table*}[t!]
\centering
\scriptsize
{\renewcommand{\arraystretch}{1.4}
\resizebox{\textwidth}{15mm}{
\begin{tabular}{ccccccccccccccccc}
\hline
\multirow{2}{*}{\textbf{FAR95}} & \multicolumn{4}{c}{\textbf{SST2}}     & \multicolumn{4}{c}{\textbf{IMDB}}     & \multicolumn{4}{c}{\textbf{TREC-10}}  & \multicolumn{4}{c}{\textbf{20NG}}     \\
                       & MSP & Energy & Maha & Cosine & MSP & Energy & Maha & Cosine & MSP & Energy & Maha & Cosine & MSP & Energy & Maha & Cosine \\ \hline
 SST2                   &  -   & -  & -  & -  & -  &  -  &  -  & -  & \textbf{5.3}/11.9    & \textbf{6.8}/10.4     & \textbf{0}/1.6    & \textbf{0.4}/6.9      & 20.5/13.7   & \textbf{4.3}/5.3       & \textbf{0}/1.2    &   \textbf{0.1}/12.6     \\
IMDB                    &  -   & -  & -  & -  & -  &  -  &  -  & -  & \textbf{0}/0.5    &\textbf{0}/0.2        & \textbf{0}/0     &\textbf{0}/0       &31.6/23.6     & 14.5/11.4       &\textbf{0.5}/4.7      &\textbf{3.5}/7.4       \\
TREC-10                &\textbf{23.2}/35.3    &\textbf{23.0}/35.0        &\textbf{0}/2.4      &\textbf{0}/4.3        &\textbf{0.6}/50.0     &\textbf{0.8}/54.0       &\textbf{0}/0      &\textbf{0}/0        &-     &-        &-      &-        &\textbf{3.5}/37.2     &\textbf{0}/13.8        &\textbf{0}/1.4      &\textbf{0}/4.4       \\
20NG                   &\textbf{15.7}/36.4    &\textbf{13.7}/36.3        &\textbf{0}/0      &\textbf{0}/0       &  \textbf{22.7}/37.8  & \textbf{24.6}/33.1       & \textbf{0}/0     & 6.3/0       & \textbf{0}/0.6    &\textbf{0}/0.2        &\textbf{0}/0      &\textbf{0}/0        &-     &-        &-      & -       \\
MNLI                   &68.7/64.6    &67.8/64.3        &7.8/0.4      &22.0/2.6      &\textbf{32.6}/52.2     &\textbf{34.8}/83.8        &\textbf{0}/0.1     &14.9/0.9        &\textbf{3.9}/9.6     & \textbf{4.5}/6.7       &\textbf{0}/0.7     &\textbf{1.5}/1.9        &38.1/37.4     &\textbf{16.9}/24.7        &\textbf{0.1}/9.6      &\textbf{0.3}/16.7        \\
RTE                    &58.5/58.3     &\textbf{57.5}/57.7        &\textbf{0}/0      &0.9/0.3       &\textbf{29.3}/52.9     &\textbf{32.6}/54.3        &\textbf{0}/0      &5.9/0.3        &\textbf{2.8}/9.8     &\textbf{3.8}/6.2        &\textbf{0}/0.1      &\textbf{0.2}/0.5        &\textbf{51.3}/52.9     &\textbf{30.6}/35.4        &\textbf{2.6}/11.1      &\textbf{4.5}/24.2        \\
WMT16                  &68.3/64.3     &67.0/64.1        &4.8/0.5      &15.3/3.0        &\textbf{18.9}/53.7     &\textbf{21.2}/55.7        &\textbf{0}/0      &2/0.4        &\textbf{5.3}/7.9     &7.2/5.7        &\textbf{0}/0.5      &\textbf{0.6}/1.3        &\textbf{37.3}/45.3     &\textbf{15.8}/27.8        & \textbf{2.0}/7.5     & \textbf{4.3}/18.7       \\
Multi30K               &44.0/36.3     &43.8/35.4        &0.2/0      &1/0.3        &\textbf{5.6}/30.9     &\textbf{5.7}/31.9        &\textbf{0}/0      &\textbf{0}/0        &\textbf{0.3}/5.3     &\textbf{0.3}/2.6       &\textbf{0}/0      &\textbf{0}/0.2       &37.9/27.8     &16.2/12.0        &\textbf{0.7}/6.9      &\textbf{3.9}/8.7       \\ \hline
Avg                    &\textbf{46.4}/49.2     &\textbf{45.5}/48.8        &2.2/0.6      &6.5/1.7       &\textbf{18.3}/46.3     & \textbf{19.9}/52.1       &\textbf{0}/0      &4.8/0.2        &\textbf{2.5}/6.5     &\textbf{3.2}/4.6        &\textbf{0}/0.4      &\textbf{0.4}/1.5        &\textbf{31.4}/32.6     &\textbf{14.0}/18.6        &\textbf{0.8}/6.0      & \textbf{2.3}/11.8  \\
\bottomrule
\end{tabular}}
\caption{OOD detection FAR95 (\%) of ours and w/ $\mathcal L_\text{margin}$ \cite{zhou-etal-2021-contrastive}. The results of our SRL-guided Self Supervision method achieving SOTA on FAR95 are highlighted in \textbf{bold}.} 
\label{table4}}
\end{table*}

\begin{table*}[!ht]
\centering
\resizebox{1.0\linewidth}{!}{
\begin{tabular}{ccc|cccc|cccc}
\hline
\multirow{2}{*}{\textbf{Baseline}} & \multirow{2}{*}{\textbf{SRL}} & \multicolumn{1}{c|}{\multirow{2}{*}{\textbf{SSL}}} & \multicolumn{4}{c|}{\textbf{IMDB } (AUROC$\uparrow$/FAR95$\downarrow$)}                                         & \multicolumn{4}{c}{\textbf{TREC }(AUROC$\uparrow$/FAR95$\downarrow$)}    \\ \cline{4-11} 
 &  &  & MSP & Energy & Maha & Cosine & MSP & Energy & Maha & Cosine \\ \hline
{\checkmark} &  &  & 93.9/46.3 & 93.4/52.1 & \textbf{1/0} & \textbf{99.9/0.2} & 97.6/6.5 & 98.1/4.6 & 99.3/0.7 & 99.0/1.5 \\
{\checkmark} & {\checkmark} &  & 95.6/25.4& 95.6/25.5 & 99.7/0.4 & 99.6/1.3 & 97.6/6.9 & 97.6/5.1 & 99.2/0.6 & 99.3/0.7 \\
{\checkmark} & {\checkmark} & {\checkmark} & \textbf{97.0/18.3} & \textbf{97.0/19.9} & \textbf{99.9/0} & 98.7/4.8 & \textbf{98.6/2.5} & \textbf{98.2/3.2} & \textbf{99.9/0} & \textbf{99.5/0.4} \\ \hline
\end{tabular}%
}
\caption{Ablation study of our method on IMDB and TREC. }
\label{abs}
\end{table*}



\subsection{Detailed Comparisons}

\label{fullresults}
In line with \cite{zhou-etal-2021-contrastive}, we also use additional datasets exclusively as OOD data, including RTE \cite{dagan2005pascal,haim2006second, giampiccolo2007third,bentivogli2009fifth}, MNLI \cite{williams-etal-2018-broad}, WMT16 \cite{bojar2016findings}, and Multi30K \cite{elliott-etal-2016-multi30k}. The comprehensive OOD detection performance on various OOD datasets is presented in Table \ref{table3} (AUC) and Table \ref{table4} (FAR). Notably, we make significant strides in improving the FAR performance by approximately 50\% when employing IMDB as ID and TREC-10 as OOD. For other OOD datasets, our approach also yields impressive FAR enhancements ranging from 10\% to 50\% when using IMDB as the ID. When employing 20NG as the ID dataset, our method significantly outperforms the previous SOTA. Collectively, our technique considerably surpasses \cite{zhou-etal-2021-contrastive}, exhibiting a substantial improvement across most benchmark measures.

\subsection{Ablation Studies and Other Analysis}

\noindent\textbf{Ablation Studies:}
In Table \ref{abs}, we substantiate the efficacy of the various proposed components on IMDB and TREC benchmarks. Please note, the baseline method refers to a model trained using $\mathcal L_\text{margin}$ without the incorporation of SRLOOD, Transformer, and Self-supervised learning modules. The results underscore that each module imparts substantial enhancements in terms of both AUROC and FAR95 on both benchmarks. This strongly affirms the effectiveness of every component proposed.

\begin{figure}[hbt]
\begin{center}
\resizebox{1\linewidth}{!}{%
\includegraphics[width=1\linewidth]{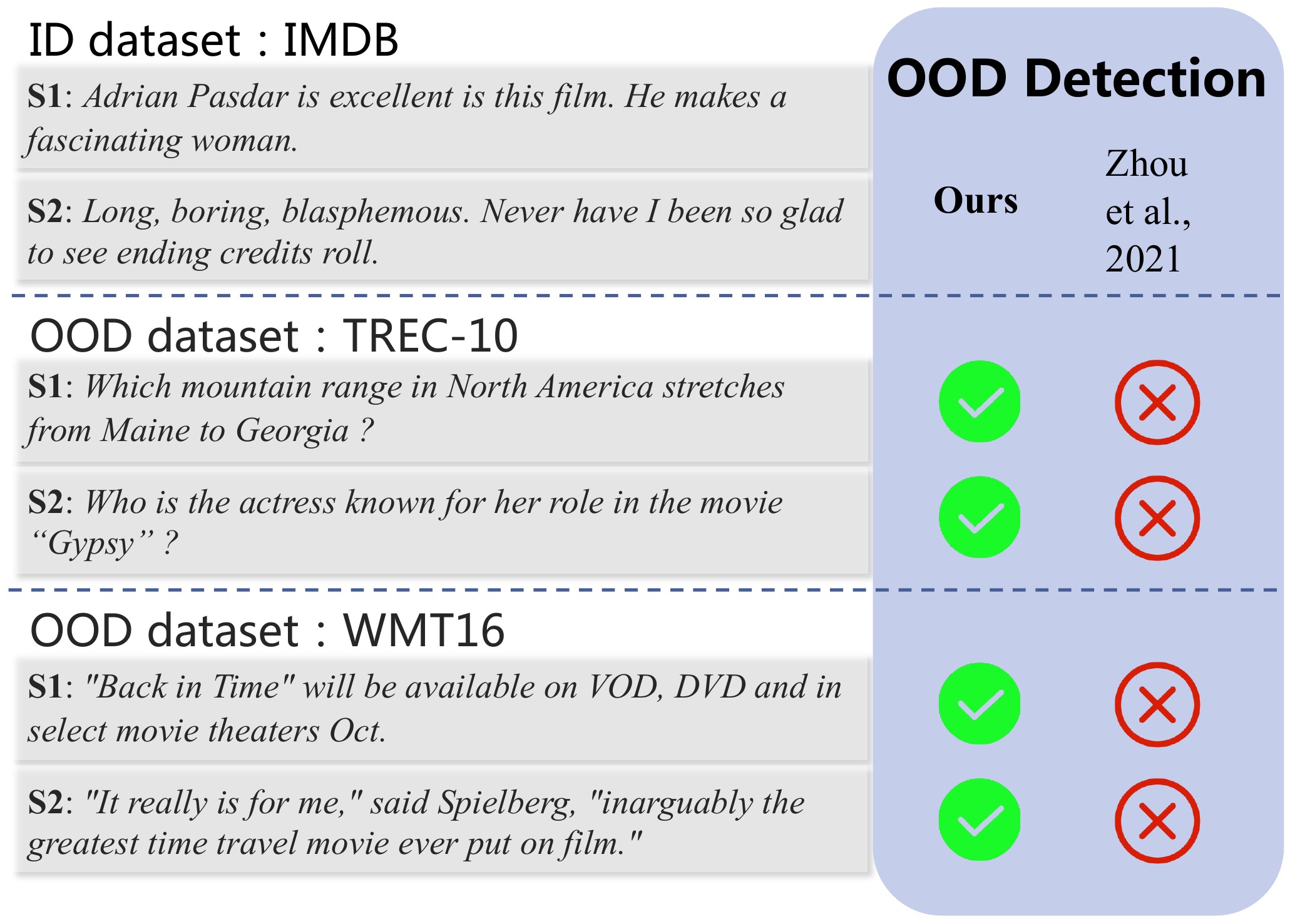}}
\end{center}

\caption{ Qualitative examples using IMDB as the In-Distribution (ID) dataset, and TREC-10 and WMT16 as the Out-of-Distribution (OOD) datasets. }
\label{fig:compare}
\end{figure}

\noindent\textbf{Qualitative Analysis:}
In Figure~\ref{fig:compare}, we present qualitative examples with IMDB functioning as the In-Distribution (ID) dataset and TREC-10 and WMT16 employed as the Out-of-Distribution (OOD) datasets. Our model exhibits superior effectiveness in identifying subtle OOD/anomalous samples, even when these contain movie content similar to the IMDB ID dataset. Conversely, the previous State-of-the-Art (SOTA) approach \cite{zhou-etal-2021-contrastive} fails to detect the majority of these sentences.

\noindent\textbf{Different masking ratios:}
\begin{figure*}[!ht] 
\begin{center}
\resizebox{1\linewidth}{!}{%
\includegraphics[width=1\linewidth]{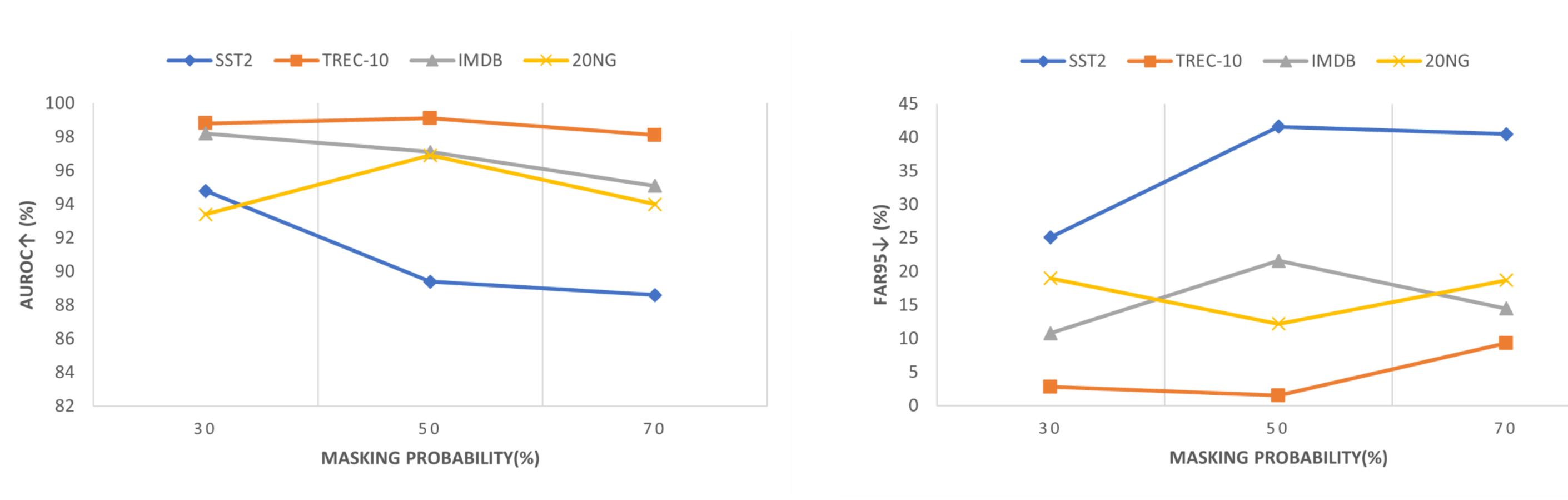}}
\end{center}
\caption{ Performance on different masking probabilities on four benchmark datasets.}
\label{maskingscore}
\end{figure*}

Figure \ref{maskingscore} illustrates three distinct masking probabilities: 0.3, 0.5, and 0.7. A higher value corresponds to stronger perturbation of the key feature representations, and we select the optimal perturbation for each ID task, guided by both the average AUROC and FAR95 over all four OOD scores, as depicted in Figure \ref{maskingscore}. The results indicate that a lower masking ratio, specifically 30\%, may yield superior performance due to its ability to retain most of the original information. On the other hand, more intense perturbation could potentially lead to a loss of crucial features. 



\section{Conclusion}

In this paper, we propose a simple yet effective approach called Semantic Role Labeling Guided Out-of-distribution Detection (SRLOOD), which learns from both global and SRL-guided local fine-grained feature representation to detect OOD instances in NLP.
The model jointly optimizes both global and local representations using a margin-based contrastive loss and self-supervised loss. 
The resulting model is able to effectively extract the key semantic roles and eliminate outliers from a sentence to detect subtle OOD samples effectively. The resulting model shows State-of-the-Art performance on four different OOD benchmarks with four different OOD scoring functions, indicating the effectiveness of our proposed SRLOOD framework. 

\nocite{*}
\section{Bibliographical References}\label{sec:reference}

\bibliographystyle{lrec-coling2024-natbib}
\bibliography{custom}


\end{document}